\documentclass{article}
\usepackage{acra}
\usepackage{fancyvrb}
\usepackage{amsmath}
\usepackage{bbm}
\usepackage{esvect}
\usepackage{amssymb}
\usepackage{amsthm}
\usepackage{verbatimbox}
\usepackage{graphicx}
\usepackage{parskip}
\usepackage[utf8]{inputenc}
\usepackage{mathtools}
\usepackage{listings}
\usepackage{breqn}
\graphicspath{ {./images/} }
\newcommand{\norm}[1]{\left\lVert#1\right\rVert}

\DeclareMathOperator*{\argmax}{arg\,max}

\usepackage{hyperref}
\usepackage{booktabs}
\usepackage[dvipsnames,table,xcdraw]{xcolor}
\usepackage[
backend=biber,
style=alphabetic,
sorting=ynt
]{biblatex}

\title{Conflict-Averse Gradient Optimization of Ensembles for Effective Offline Model-Based Optimization}
\author{Sathvik Kolli \\ UC Berkeley Electrical Engineering \& Computer Science \\  sathkolli@berkeley.edu}

\begin{document}

\maketitle

\begin{abstract}
Data-driven offline model-based optimization (MBO) is an established practical approach to black-box computational design problems for which the true objective function is unknown and expensive to query. However, the standard approach which optimizes designs against a learned proxy model of the ground truth objective can suffer from distributional shift. Specifically, in high-dimensional design spaces where valid designs lie on a narrow manifold, the standard approach is susceptible to producing out-of-distribution, invalid designs that ``fool" the learned proxy model into outputting a high value. Using an ensemble rather than a single model as the learned proxy can help mitigate distribution shift, but naive formulations for combining gradient information from the ensemble, such as minimum or mean gradient, are still suboptimal and often hampered by non-convergent behavior. 

In this work, we explore alternate approaches for combining gradient information from the ensemble that are robust to distribution shift without compromising optimality of the produced designs. More specifically, we explore two functions, formulated as convex optimization problems, for combining gradient information: multiple gradient descent algorithm (MGDA) \cite{mgda} and conflict-averse gradient descent (CAGrad) \cite{conflictaverse}. We evaluate these algorithms on a diverse set of five computational design tasks \cite{designbench}. We compare performance of ensemble MBO with MGDA and ensemble MBO with CAGrad with three naive baseline algorithms: (a) standard single-model MBO, (b) ensemble MBO with mean gradient, and (c) ensemble MBO with minimum gradient.

Each algorithm produces 128 optimized designs, and we report performance of these designs under three metrics: (a) max ground truth score, (b) average ground truth score, (c) 50th percentile ground truth score. 
\begin{itemize}
    \item For the max ground truth score, we find that MGDA is in the top 2 best-performing algorithms on 4 tasks and is the best-performing algorithm on 2 tasks. CAGrad is the in the top 2 best-performing algorithms on 2 tasks.
    \item For the 50th percentile ground truth score, MGDA is the best-performing algorithm on 1 task. CAGrad is in the top 2 best-performing algorithms on 3 tasks and is the best-performing algorithm on 2 tasks.
    \item Finally, for the average ground truth score, MGDA is the best-performing algorithm on 2 tasks. CAGrad is in the top 2 best-performing algorithms on 3 tasks and is the best-performing algorithm on 2 tasks.
\end{itemize}

Our results suggest that MGDA and CAGrad strike a desirable balance between conservatism and optimality. In general, we noticed that MGDA and CAGrad performed equally well, if not better, than other algorithms on the max ground truth score. However, both algorithms lead to significant improvement on the average and 50th percentile ground truth scores, suggesting they may be more conservative and less susceptible to being ``fooled" by invalid designs. Our results demonstrate that MGDA and CAGrad can help robustify data-driven offline MBO without compromising optimality of designs.
\end{abstract}

\vspace{5mm}
\section{Introduction}

We study the problem of computational design, which arises in settings ranging from synthetic biology to robot design. Specifically, we focus on the setting of black-box optimization, which attempts to generate optimal designs where the objective function and constraints are unknown. Put simply, we want to find the optimal design, $x$, that maximizes some unknown objective function, $f(x)$:
$$\argmax_x \enspace f(x)$$
Examples of black-box optimization problems include optimizing robot morphologies, biological sequences (proteins, genes), computer chips, neural network architectures, or superconducting materials.  

\vspace{3mm}
\subsection{Offline Model Based Optimization (MBO)}
One promising approach to solving black-box optimization problems is data-driven MBO, where a proxy model of the unknown objective function is learned from empirically collected data and used to guide the design procedure. 

In order to model the true objective function with high fidelity, it is often critical to actively collect additional data during the training procedure \cite{online}. However, in many design problems, active real-world data collection is expensive (e.g. requires synthesizing protein structures for protein optimization or building and testing a robot for robot design) or dangerous (e.g. when optimizing over aircraft designs). Thus, we focus instead on the more practical setting of offline MBO, where we are given a static dataset of designs and cannot make any queries to the ground truth.

In essence, when we use offline MBO to solve black-box optimization problems, we are trying to solve the problem
$$\argmax_x \enspace f(x)$$
with two key assumptions:
\begin{enumerate}
    \item Black-box assumption: $f(x)$ is an unknown function
    \item Offline assumption: $f(x)$ is expensive to query
\end{enumerate}

\begin{figure}
    \centering
    \includegraphics[width=\columnwidth]{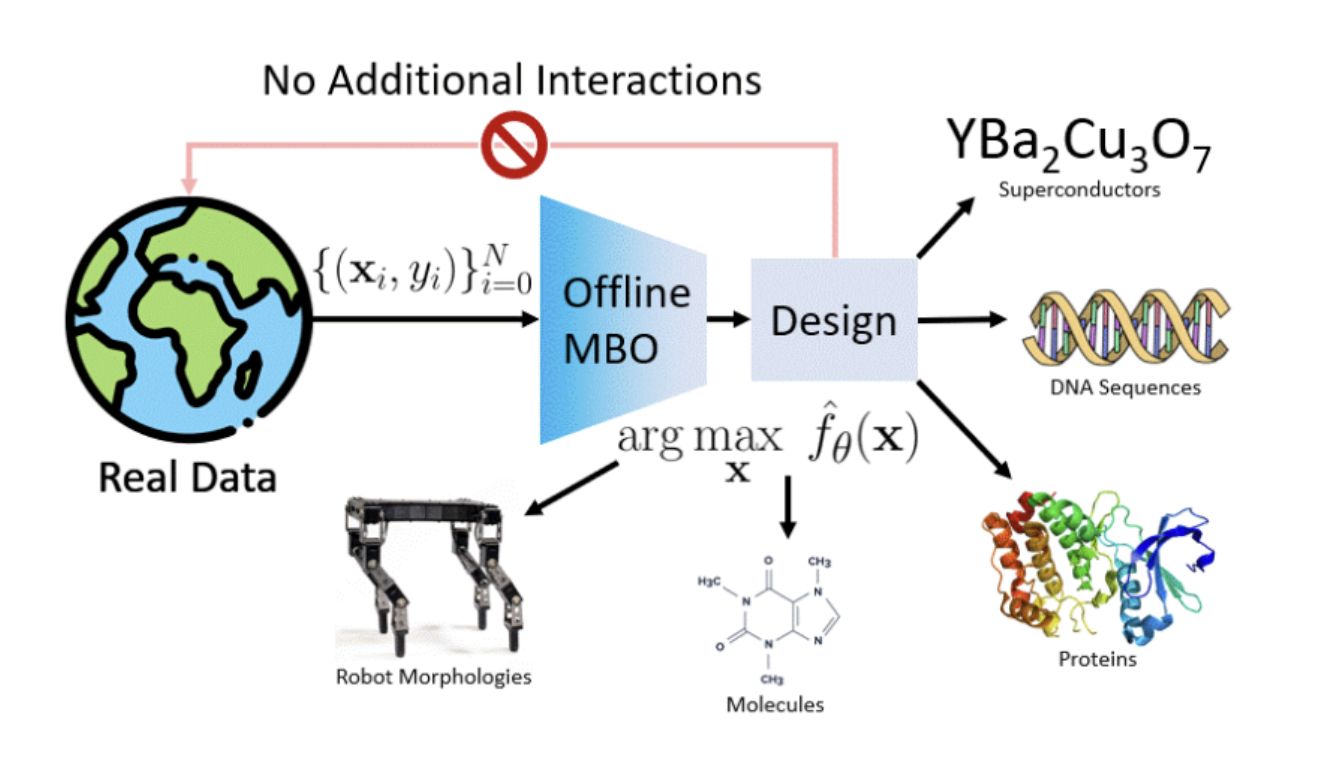}
    \caption{A typical offline MBO workflow \cite{designbench}. We are given a static dataset of designs, which we use to learn a proxy model, $\hat{f}_\theta$, of the true objective. Then, our design procedure is guided by the learned proxy model.}
    \label{fig:offmbo}
\end{figure}

The general offline MBO workflow is illustrated in Figure \ref{fig:offmbo}. 

\vspace{3mm}
\subsection{Distribution Shift in MBO}

The most basic approach to offline, data-driven model-based optimization involves the following steps \cite{coms}:
\begin{enumerate}
	\item We have a static dataset $D$ of input designs and their corresponding objective values: $$\{(x_1, y_1), \hdots, (x_N, y_N)\}$$ We assume this paired data was generated from a true, unknown objective function, $y = f(x)$.
	\item Using the dataset $D$, we learn a proxy model $\hat{f}_\theta(x)$ of the unknown objective function $f(x)$, via supervised regression on the training dataset.
	\item Finally, we find an optimal generated design $x^*$, by optimizing data point $x \in D$ against the learned model $\hat{f}_\theta(x)$ (e.g. using $T$ gradient ascent/descent steps on the learned function):
	$$x_{k+1} \leftarrow x_k + \alpha \nabla_x f_\theta(x)|_{x = x_k}, \text{ for } k \in [1, T]$$
\end{enumerate}

As described, the above approach does not perform well in high-dimensional input spaces, where the space of valid input designs lie on a narrow manifold, because overestimation errors in the proxy model $\hat{f}_\theta(x)$ would cause to the optimization procedure in step (3) to produce out-of-distribution, invalid, and low-valued designs. Consequently, for the above method to work, it is critical that we ensure that the proxy model, $f_\theta(x)$, does not overestimate the objective value of out-of-distribution points. 

Some existing approaches to prevent overestimation of out-of-distribution inputs include generative modeling, explicit density estimation, or regularization techniques that incentivize conservatism in regions with limited data (known as Conservative Objective Models \cite{coms}).

\vspace{3mm}
\subsection{Ensemble MBO}

One simple approach to addressing the issue of distribution shift is to use an ensemble rather than a single model as a proxy for the true objective function. The rationale behind this is that it is less likely for multiple different models to be ``fooled” by the same out-of-distribution input than it is for a single model to suffer from overestimation errors.

The ensemble can have individual models with varied architectures and regularization techniques. Although we don't study this in this project, future work can try including density models and/or Conservative Objective Models with varying levels of conservatism in the ensemble.

When we used a single proxy model, we used gradient ascent on the model in order to generate new designs:
$$x_{k+1} \leftarrow x_k + \alpha \nabla_x \hat{f}_\theta(x_k)$$
Now, that we use an ensemble $\{\hat{f}_1(x), \hdots, \hat{f}_m(x)\}$, we need to update our design using gradient information from all the models in the ensemble. Thus, we get the following update:
$$x_{k+1} \leftarrow x_k + \alpha g(\nabla_x \hat{f}_1(x_k), \hdots, \nabla_x \hat{f}_m(x_k))$$
where $g$ is some function of all the gradients.

Two naive approaches for the function $g$ are:
\begin{itemize}
    \item Mean Gradient: $$\nabla_x \left(\sum_{i = 1}^{m} f_i(x)\right)$$
    The benefit of this approach is that it captures gradient information from all the models in the ensemble in each gradient step. 
    
    However, while this approach may lead to optimization that is more robust to distribution shift, it is still possible for the optimization to be ``fooled" by an out-of-distribution input, particularly when a model or group of models within the ensemble dominate the update. Furthermore, it is also possible for the optimization to get stuck and fail to optimize further (sometimes becoming stuck in a oscillating manner) due to conflicting gradients.
    \item Minimum Gradient: $$\nabla_x \min\left(f_1(x), f_2(x), \hdots, f_m(x)\right)$$
    The benefit of this approach is that it is conservative and is less likely to be ``fooled" by an out-of-distribution input. In fact, we can interpret the minimum gradient update as searching for a design for which \textit{all} the models in the ensemble assign a high score. By definition, this algorithm would produce an out-of-distribution point if and only if every model in the ensemble overestimated the value of that point.

    The drawback of this method is that it has poor convergence guarantees and is susceptible to oscillatory behavior.
\end{itemize}

\vspace{3mm}
\subsection{Multiple Gradient Descent Algorithm (MGDA) and Conflict-Averse Gradient Descent (CAGrad)}

We consider two alternative functions for $g$ that combine the gradient information from the models in the ensemble: multiple gradient descent algorithm (MGDA) \cite{mgda} and conflict-averse gradient descent (CAGrad) \cite{conflictaverse}. Both functions are formulated as convex optimization problems. MGDA was developed in the setting of multi-objective optimization, while CAGrad was developed for multi-task learning. 

For MGDA, the gradient is defined in terms of the following convex optimization problem:
$$\max_{d} \enspace \min_i \enspace \langle d, g_i \rangle - \frac{1}{2}\norm{d}^2$$
where 
$$g_i = \nabla_x \hat{f}_i(x), \enspace i = 1, \hdots, m$$
For CAGrad, the gradient is defined in terms of the following convex optimization problem:
$$\max_{d} \enspace \min_i \enspace \langle d, g_i \rangle \enspace \text{s.t.} \enspace \norm{d - g_0} \leq c\norm{g_0}$$
where 
$$g_i = \nabla_x \hat{f}_i(x), \enspace i = 1, \hdots, m$$ 
$c \in [0, 1)$ is a hyper-parameter, 
and $g_0 = \frac{1}{m} \nabla_x \hat{f}_i(x)$ is the average gradient. 

The high-level intuition behind these methods is that they search for a design with a high model-predicted objective value, while leveraging the worst local improvement of the models in the ensemble to regularize the optimization trajectory. 

More specifically, assume we update our design $x$ by $x \leftarrow x + \alpha d$, where $\alpha$ is a step size and $d$ is the update vector. Then, the minimum improvement rate across the models in the ensemble is given by:
\begin{align*}
    R(x, d) &= \min_{i \in \{1, \hdots, m\}} \left(\frac{1}{\alpha}(f_i(x + \alpha d) - f_i(x))\right) \\
    &\approx \min_{i \in \{1, \hdots, m\}} \langle g_i, d \rangle
\end{align*}
where we use the first-order Taylor approximation, assuming $\alpha$ is small. 

We can view both MGDA and CAGrad as looking for the ``best" update vector $d$ within a local ball, where we define the ``best" update vector as the one that maximizes the worst improvement rate. The difference between the two is the local ball within which we search for the optimal update vector, where MGDA is centered at zero, while CAGrad is centered at the average gradient $g_0$.

\begin{figure}
    \centering
    \includegraphics[scale=1.75]{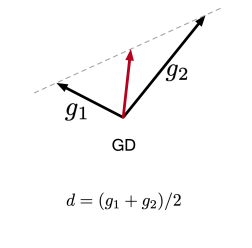}
    \includegraphics[scale=1.75]{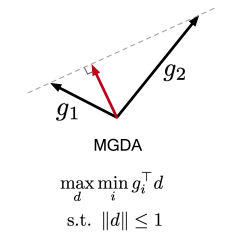}
    \includegraphics[scale=1.75]{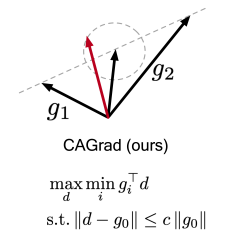}
    \caption{Comparison of methods from ``Conflict-Averse Gradient Descent for Multi-task Learning'' \cite{conflictaverse}. We compare naive gradient descent (GD, top) to MGDA (middle) and CAGrad (bottom).}
    \label{fig:grads}
\end{figure}

In theory, MGDA provably converges to an arbitrary point on the Pareto set \cite{mgda}, while CAGrad provably converges to a stationary point of the average proxy objective, when $0 \leq c < 1$ \cite{conflictaverse}.

The differences between mean gradient, MGDA, and CAGrad are illustrated in Figure \ref{fig:grads}, taken from \cite{conflictaverse}.

\vspace{5mm}
\section{Methods}

We evaluate the performance of five algorithms
\begin{itemize}
    \item Single Proxy Model, Naive Gradient Ascent
    \item Ensemble Proxy Model, Mean Gradient Ascent
    \item Ensemble Proxy Model, Min Gradient Ascent
    \item Ensemble Proxy Model, MGDA
    \item Ensemble Proxy Model, CAGrad
\end{itemize}
on five diverse benchmark tasks selected from \cite{designbench}. The tasks and relevant details are listed in Table \ref{tab:comparisons}.

\begin{table*}[t]
  \centering
  \begin{tabular}{@{}lllllll@{}}
\toprule
                           & \textbf{\begin{tabular}[c]{@{}l@{}}Total\\ Dataset\\ Size\end{tabular}} & \textbf{\begin{tabular}[c]{@{}l@{}}MBO\\ Dataset\\ Size\end{tabular}} & \textbf{Dimensions} & \textbf{Type} & \textbf{\begin{tabular}[c]{@{}l@{}}Oracle (Spear.\\ Correlation)\end{tabular}} & \textbf{\begin{tabular}[c]{@{}l@{}}Primal or \\ Dual? (\# of \\decision var.) \end{tabular}} \\ \midrule
\textbf{TF Bind 8}         & 65536                                                                   & 32898                                                                 & (8, 4)              & Discrete      & Lookup Table          & Primal (32)                                                        \\
\textbf{TF Bind 10}        & 1048576                                                                 & 50000                                                                 & (10, 4)             & Discrete      & Lookup Table          & Primal (40)                                                        \\
\textbf{ChEMBL}            & 1093                                                                    & 546                                                                   & (31, 591)           & Discrete      & Random Forest (0.792) & Dual (6)                                                           \\
\textbf{Hopper Controller} & 3200                                                                    & 3200                                                                  & 5126                & Continuous    & Exact                 & Dual (6)                                                           \\
\textbf{Ant Morphology}    & 25009                                                                   & 15005                                                                 & 60                  & Continuous    & Exact                 & Primal (60)                                                           \\ \bottomrule
\end{tabular}
\begin{tabular}{@{}lllll@{}}
\toprule
                           & \textbf{\begin{tabular}[c]{@{}l@{}}Grad. Asc.\\ Parameters \\ (\# of rounds, $\alpha$)\end{tabular}} & \textbf{\begin{tabular}[c]{@{}l@{}}Design Model\\ Architecture \\ (\# of parameters)\end{tabular}} & \textbf{\begin{tabular}[c]{@{}l@{}}Design Model\\ Results (Val. Spear., \\ Val. Loss)\end{tabular}} & \textbf{\begin{tabular}[c]{@{}l@{}}CAGrad\\ Parameter\end{tabular}} \\ \midrule
\textbf{TF Bind 8}         & (200, 10)                                                             & FullyConnected (1140801)                                                     & (0.43,  0.15)                                                                         & c=0.5                                                            \\
\textbf{TF Bind 10}        & (200, 50)                                                             & FullyConnected (2122753)                                                     & (0.97, 0.00)                                                                          & c=0.5                                                            \\
\textbf{ChEMBL}            & (200, 100)                                                            & FullyConnected (6582401)                                                     & (0.77, 0.13)                                                                          & c=0.5                                                            \\
\textbf{Hopper Controller} & (200, 1)                                                                 & FullyConnected (5252097)                                                     & (0.87, 0.28)                                                                          & c=0.3                                                            \\
\textbf{Ant Morphology}    & (200, 0.03)                                                              & FullyConnected (8460289)                                                     & (0.57, 0.33)                                                                          & c=0.2                                                            \\ \bottomrule
\end{tabular}
\caption{Comparison of the TF Bind 8, TF Bind 10, ChEMBL, Hopper, and Ant Morphology tasks.}
  \label{tab:comparisons}
\end{table*}

\vspace{3mm}
\subsection{General Procedure}

Each task has a corresponding dataset, which we refer to as the ``total dataset" for the task. In order to evaluate our MBO algorithms, we take a subset of the total dataset, which we call the ``MBO dataset." 

The general procedure we use for evaluation is as follows:
\begin{enumerate}
    \item The design/proxy model(s) are trained on the MBO dataset, which is a subset (bottom K\% of target values) of the total dataset for the task. 
    \item We take the top 128 (i.e. highest target values) inputs in the MBO dataset and optimize each one using our algorithm to produce 128 designs.
    \item Next, we calculate the (a) max ground-truth score, (b) the 50th percentile ground-truth score, and (c) the average ground-truth score of the 128 designs. 
\end{enumerate}

For all the ensemble proxy model methods, we use six ensemble models, which all have the same architecture, but are trained and validated on different subsets of the MBO dataset.

The motivation for the above approach is that, in real design scenarios, it is usually impractical to empirically test and use every design produced by a given algorithm. Instead, we would only select the top-scored designs according to our algorithm and experimentally validate them and use them. Thus, ideally, we want an algorithm whose best set of designs is actually the best under the true objective function.

\vspace{3mm}
\subsection{Oracles}

For the third step in the above procedure, some tasks have an exact oracle, meaning that the ground truth for every possible permutation of inputs is provided in a lookup table or that individual designs are cheap to evaluate, while other tasks use a learned oracle (e.g. neural network, random forest model) as a proxy for the ground truth. In this case, we train a separate oracle model on the total dataset for the task. Details on the oracles for each task are listed in Table \ref{tab:comparisons}.

\vspace{3mm}
\subsection{Dual Formulations}

When we use the primal formulation of MGDA and CAGrad, which are presented above, the number of decision variables is equal to the dimensionality of the input designs. In some cases, this is computationally feasible. However, for some of our tasks, the inputs are high-dimensional, and we need to use the dual formulation of MGDA and CAGrad, where the number of decision variables is equal to the number of models in the ensemble (i.e. 6). For both MGDA and CAGrad, since the primal problem is convex and Slater's condition holds, we have strong duality.

As before, we define:
$$g_i = \nabla_x \hat{f}_i(x), \enspace i = 1, \hdots, m$$ $c \in [0, 1)$ is a hyper-parameter for CAGrad, 
and $g_0 = \frac{1}{m} \nabla_x \hat{f}_i(x)$ is the average gradient. 

The dual formulation of MGDA is
$$ \min_w \frac{1}{2} \norm{\sum_{i=1}^K w_i g_i }^2 \enspace \text{s.t.} \enspace \sum_{i=1}^K w_i = 1 \enspace \text{and} \enspace \forall i, w_i \geq 0$$
\cite{conflictaverse}.

The dual formulation of CAGrad is
$$\min_w g_w^Tg_0 + \sqrt{\phi}||g_w|| \enspace \text{s.t.} \enspace \sum_{i=1}^K w_i = 1 \enspace \text{and} \enspace \forall i, w_i \geq 0 $$
where $g_w = \sum_iw_ig_i$ and $\phi = c^2||g_0||^2$. The optimal update vector is $d^* = g_0 + g_{w^*} / \lambda^*$, where $\lambda^* = \norm{g_{w^*}} / \sqrt{\phi}$,  \cite{conflictaverse}.

We can interpret the dual formulations as finding the weights for a weighted average of the gradients from each model in the ensemble.

Details regarding which formulation we use for each benchmark task are listed in Table \ref{tab:comparisons}.

\vspace{3mm}
\subsection{Gradient Ascent Procedure}

For discrete tasks, we perform gradient updates in one-hot space. After each update, we map the sequence back to discrete space by taking an argmax.

For continuous tasks, we normalize the inputs, so that each position has zero mean and unit variance. We then perform gradient updates in this space. 

\vspace{3mm}
\subsection{Hyperparameter Selection}

Each task has three relevant hyperparameters:
\begin{itemize}
    \item Design Model Architecture and Training Parameters
    \item Gradient Ascent Parameters: (Number of Gradient Update Steps, Learning Rate)
    \item CAGrad Hyperparameter $c$
\end{itemize}

In order to select and train the design model architecture, we randomly split the MBO Dataset into a training set and validation set, and we selected the model with the best validation spearman correlation and validation mean-squared loss.

For the gradient ascent parameters, we fixed the number of gradient update steps to 200, and we selected the learning rate by visually analyzing optimization trajectories (i.e. plot of proxy model(s) prediction vs. number of rounds of mutation). We selected the CAGrad hyperparameter $c$ in a similar fashion.

It is important that hyperparameter tuning should be done purely offline, without any access to the ground truth objective or oracle. 

Details regarding the hyperparameters we use for each task are listed in Table \ref{tab:comparisons}.

\vspace{3mm}
\subsection{Discrete Benchmark Tasks}

We detail the three discrete tasks on which we perform evaluation. 

The \textbf{TF Bind 8} task \cite{tfbind8} is based on an empirical dataset of measurements of binding activity between a variety of human transcription factors and every possible length-8 DNA sequence. The optimization goal is to identify DNA sequences that maximizing the binding activity score for each TF. The design space for sequences is comprised of four categorical variables, one representing each nucleotide (A, T, C, or G). The oracle for TF Bind 8 is exact (a lookup table containing ground-truth values for every possible permutation of inputs).

The \textbf{TF Bind 10} task \cite{tfbind10} is a neural network produced dataset of predicted estimates of the relative binding affinities between all unique length-10 DNA sequences and each of two protein targets. The optimization goal is to identify DNA sequences that maximize the predicted binding affinity to targets. The design space for sequences is A, T, C, and G as before. The oracle for TF Bind 10 is exact (a lookup table containing ground-truth values for every possible permutation of inputs).

The \textbf{ChEMBL} task \cite{chembl} is a dataset of pairs of molecules and assays which test for specific functional properties of those molecules. The optimization goal is to design a molecule that achieves a high functional property score on a specific assay. The design space for molecules is based on SMILES encodings (rather than amino acids), resulting in a design space of categorical variables that take one of 591 values, for sequences of length 31.  The oracle for ChEMBL is a random forest, which achieves a spearman correlation of 0.792 on the dataset.

\vspace{3mm}
\subsection{Continuous Benchmark Tasks}

We detail the two continuous tasks on which we perform evaluation. 

The \textbf{Hopper Controller} task is an OpenAI gym locomotion task \cite{gym}. The optimization goal is to design a set of weights for a controller neural network (representing a policy) that will optimize for expected return on the locomotion task. While Hopper is typically a reinforcement learning task, we formulate it as offline MBO by utilizing a supervised dataset of neural network controlled weights matched with expected return values. There are 5126 continuous variables corresponding to the flattened weights of this neural network. In order to evaluate the ground truth score for a design, we simply load in the weights of the neural network controller and run 1000 steps of simulation in the MuJoCo simulator \cite{mujoco} used with this environment.

The \textbf{Ant Morphology} task is an OpenAI gym task \cite{gym}. The goal is to optimize the morphology (e.g. size, orientation, location of limbs) of Ant, a simulated robot whose goal is to run fast (i.e. a locomotion task) in its environment. There are 60 continuous variables corresponding to these morphological parameters. We obtain a design's ground truth score by running robotic simulation in the MuJoCo simulator \cite{mujoco} for 100 time steps, averaging 16 independent trials.

\vspace{5mm}
\begin{table*}[t]
  \centering
\begin{tabular}{|l|l|l|l|l|l|}
\hline
                          & \textbf{TF Bind 8}                     & \textbf{TF Bind 10}                    & \textbf{\begin{tabular}[c]{@{}l@{}}ChEMBL\\ (Random Forest Oracle)\end{tabular}} & \textbf{\begin{tabular}[c]{@{}l@{}}Hopper\\ Controller\end{tabular}} & \textbf{\begin{tabular}[c]{@{}l@{}}Ant\\ Morphology\end{tabular}} \\ \hline
\rowcolor[HTML]{FFF2CC} 
\textbf{dataset}          & \textbf{0.439}                         & \textbf{0.240}                         & \textbf{0.635}                                                                   & \textbf{1.0}                                                         & \textbf{0.747}                                                    \\ \hline
\textbf{single model}     & \cellcolor[HTML]{6D9EEB}\textbf{0.976} & \textbf{0.682}                         & \cellcolor[HTML]{93C47D}\textbf{0.808}                                           & \textbf{1.544}                                                       & \textbf{0.807}                                                    \\ \hline
\textbf{ensemble, mean}   & \textbf{0.973}                         & \cellcolor[HTML]{93C47D}\textbf{0.754} & \textbf{0.777}                                                                   & \textbf{2.829}                                                       & \textbf{0.944}                                                    \\ \hline
\textbf{ensemble, min}    & \cellcolor[HTML]{6D9EEB}\textbf{0.976} & \textbf{0.726}                         & \textbf{0.788}                                                                   & \cellcolor[HTML]{6D9EEB}\textbf{3.040}                               & \cellcolor[HTML]{93C47D}\textbf{0.977}                            \\ \hline
\textbf{ensemble, MGDA}   & \cellcolor[HTML]{93C47D}\textbf{0.979} & \textbf{0.734}                         & \cellcolor[HTML]{6D9EEB}\textbf{0.800}                                           & \cellcolor[HTML]{93C47D}\textbf{3.579}                               & \cellcolor[HTML]{6D9EEB}\textbf{0.949}                            \\ \hline
\textbf{ensemble, CAGrad} & \cellcolor[HTML]{6D9EEB}\textbf{0.976} & \cellcolor[HTML]{6D9EEB}\textbf{0.735} & \textbf{0.774}                                                                   & \textbf{2.815}                                                       & \textbf{0.924}                                                    \\ \hline
\end{tabular}
  \caption{\textbf{Max (Normalized) ground-truth score of the top 128 generated designs.} For each task, the best score is green, and the second best score is blue. The ``dataset" row in yellow is the normalized score of the best design in the starting offline MBO dataset.}
  \label{tab:2}
\end{table*}

\begin{table*}[t]
  \centering
\begin{tabular}{|
>{\columncolor[HTML]{FFFFFF}}l |
>{\columncolor[HTML]{FFFFFF}}l |
>{\columncolor[HTML]{FFFFFF}}l |
>{\columncolor[HTML]{FFFFFF}}l |
>{\columncolor[HTML]{FFFFFF}}l |
>{\columncolor[HTML]{FFFFFF}}l |}
\hline
                          & \textbf{TF Bind 8}                     & \textbf{TF Bind 10}                    & \textbf{\begin{tabular}[c]{@{}l@{}}ChEMBL\\ (Random Forest Oracle)\end{tabular}} & \textbf{\begin{tabular}[c]{@{}l@{}}Hopper\\ Controller\end{tabular}} & \textbf{\begin{tabular}[c]{@{}l@{}}Ant\\ Morphology\end{tabular}} \\ \hline
\textbf{single model}     & \cellcolor[HTML]{6D9EEB}\textbf{0.758} & \textbf{0.568}                         & \textbf{0.740}                                                                   & \cellcolor[HTML]{93C47D}\textbf{0.658}                               & \textbf{0.415}                                                    \\ \hline
\textbf{ensemble, mean}   & \textbf{0.750}                         & \textbf{0.685}                         & \cellcolor[HTML]{93C47D}\textbf{0.770}                                           & \cellcolor[HTML]{6D9EEB}\textbf{0.655}                               & \textbf{0.689}                                                    \\ \hline
\textbf{ensemble, min}    & \textbf{0.737}                         & \cellcolor[HTML]{6D9EEB}\textbf{0.688} & \textbf{0.685}                                                                   & \textbf{0.646}                                                       & \cellcolor[HTML]{6D9EEB}\textbf{0.707}                            \\ \hline
\textbf{ensemble, MGDA}   & \textbf{0.683}                         & \textbf{0.686}                         & \textbf{0.760}                                                                   & \textbf{0.650}                                                       & \cellcolor[HTML]{93C47D}\textbf{0.735}                            \\ \hline
\textbf{ensemble, CAGrad} & \cellcolor[HTML]{93C47D}\textbf{0.811} & \cellcolor[HTML]{93C47D}\textbf{0.692} & \cellcolor[HTML]{6D9EEB}\textbf{0.768}                                           & \textbf{0.629}                                                       & \textbf{0.704}                                                    \\ \hline
\end{tabular}
  \caption{\textbf{50th Percentile (Normalized) ground-truth score of the top 128 generated designs.} For each task, the best score is green, and the second best score is blue.}
  \label{tab:3}
\end{table*}

\begin{table*}[t]
  \centering
\begin{tabular}{|
>{\columncolor[HTML]{FFFFFF}}l |
>{\columncolor[HTML]{FFFFFF}}l |
>{\columncolor[HTML]{FFFFFF}}l |
>{\columncolor[HTML]{FFFFFF}}l |
>{\columncolor[HTML]{FFFFFF}}l |
>{\columncolor[HTML]{FFFFFF}}l |}
\hline
                          & \textbf{TF Bind 8}                     & \textbf{TF Bind 10}                    & \textbf{\begin{tabular}[c]{@{}l@{}}ChEMBL\\ (Random Forest Oracle)\end{tabular}} & \textbf{\begin{tabular}[c]{@{}l@{}}Hopper\\ Controller\end{tabular}} & \textbf{\begin{tabular}[c]{@{}l@{}}Ant\\ Morphology\end{tabular}} \\ \hline
\textbf{single model}     & \cellcolor[HTML]{6D9EEB}\textbf{1.524} & \textbf{0.404}                         & \textbf{0.547}                                                                   & \textbf{563.45}                                                      & \textbf{19.426}                                                   \\ \hline
\textbf{ensemble, mean}   & \textbf{1.494}                         & \textbf{1.112}                         & \cellcolor[HTML]{93C47D}\textbf{0.706}                                           & \textbf{578.11}                                                      & \textbf{251.980}                                                  \\ \hline
\textbf{ensemble, min}    & \textbf{1.410}                         & \cellcolor[HTML]{6D9EEB}\textbf{1.138} & \textbf{0.285}                                                                   & \cellcolor[HTML]{6D9EEB}\textbf{598.55}                              & \cellcolor[HTML]{6D9EEB}\textbf{281.266}                          \\ \hline
\textbf{ensemble, MGDA}   & \textbf{1.055}                         & \textbf{1.117}                         & \textbf{0.615}                                                                   & \cellcolor[HTML]{93C47D}\textbf{602.12}                              & \cellcolor[HTML]{93C47D}\textbf{306.254}                          \\ \hline
\textbf{ensemble, CAGrad} & \cellcolor[HTML]{93C47D}\textbf{1.895} & \cellcolor[HTML]{93C47D}\textbf{1.150} & \cellcolor[HTML]{6D9EEB}\textbf{0.695}                                           & \textbf{564.72}                                                      & \textbf{236.802}                                                  \\ \hline
\end{tabular}
  \caption{\textbf{Average (Unnormalized) ground-truth score of the top 128 generated designs.} For each task, the best score is green, and the second best score is blue.}
  \label{tab:4}
\end{table*}

\section{Results}

We report three metrics from the top 128 designs of each algorithm: (a) max ground truth score (Table \ref{tab:2}), (b) 50th percentile ground truth score (Table \ref{tab:3}), (c) average ground truth score (Table \ref{tab:4}).

In order to report performance on the same order of magnitude across tasks, we normalize the max ground truth score and the 50th percentile ground truth scores using the formula
$$y_{\text{normalized}}(y) = \frac{y - y_{\min}}{y_{\max} - y_{\min}}$$
where $y_{\max}$ and $y_{\min}$ are the maximum and minimum objective values in the total dataset for each task. By definition, a normalized score greater than 1 means that we have designed an input that is better than any input in the total dataset for the task.

We do not normalize the average ground truth scores.

Here are the results we observed:
\begin{itemize}
    \item For the max ground truth score, we find that MGDA is in the top 2 best-performing algorithms on 4 tasks and is the best-performing algorithm on 2 tasks. CAGrad is the in the top 2 best-performing algorithms on 2 tasks.
    \item For the 50th percentile ground truth score, MGDA is the best-performing algorithm on 1 task. CAGrad is in the top 2 best-performing algorithms on 3 tasks and is the best-performing algorithm on 2 tasks.
    \item Finally, for the average ground truth score, MGDA is the best-performing algorithm on 2 tasks. CAGrad is in the top 2 best-performing algorithms on 3 tasks and is the best-performing algorithm on 2 tasks.
\end{itemize}

In general, we observe that MGDA and CAGrad performed roughly as well, if not better, than other algorithms on the max ground truth score. However, when considering the 50th percentile and average ground truth scores, we found that MGDA performed much better than other algorithms on the continuous tasks and CAGrad performed much better than other algorithms on the discrete tasks. This suggests that that MGDA and CAGrad are more conservative and less susceptible to being ``fooled" by invalid designs.

\vspace{5mm}
\section{Discussion and Future Work}

\vspace{3mm}
\subsection{Interpretation of Results}

Based on our results, MGDA and CAGrad seem to robustify data-driven offline MBO without compromising optimality of designs. In real-world design scenarios, utilizing MGDA or CAGrad over mean or minimum gradient could be well-motivated in contexts where we care about generating a diverse dataset of good designs rather than one-off good designs. Often, computational design projects involve repeated iteration between design generation and experimental validation. Usually, it is more practical and efficient to experimentally validate many proposed good designs at a time (e.g. a dataset), rather than repeated iteration over a single design.

\vspace{3mm}
\subsection{Future Work}

We faced some key challenges in this work that present opportunities for future research. First, hyperparameter selection in a purely offline manner is difficult, and future work should explore better, more rigorous methods for offline hyperparameter tuning. 

Second, there are a lot of different approaches for using a gradient ascent optimizer in discrete space: gradient normalization, alternating between updates in soft-space and updates in hard-space, and more. In our experimentation, we found that certain methods, such as gradient normalization, improved the performance of MGDA on discrete tasks significantly, but for consistency, we present results for a relatively simple gradient ascent optimizer. Future work should study alternate methods for gradient ascent in discrete space. 

Finally, a key challenge we faced was with the ChEMBL task and other tasks we tried out which don't have an exact oracle. Using a learned oracle to evaluate how robust our design algorithms are to out-of-distribution designs is unreliable, because learned oracles usually suffer from the same distribution shift problem as the proxy design models. Finding a way to reliably evaluate design algorithms using learned oracles is an important area for future research, because many real-world design tasks don't have exact oracles.

Finally, although we use a diverse set of tasks, future work can study CAGrad and MGDA on more tasks, especially with a focus on the unique characteristics of tasks that may make MGDA more suitable than CAGrad, or vice versa. 

\newpage
\cleardoublepage

\vspace{5mm}
\printbibliography[
heading=bibintoc,
title={References}
]

\end{document}